\documentclass[runningheads]{llncs}
\usepackage[T1]{fontenc}
\usepackage{graphicx}
\usepackage{adjustbox}
\usepackage{tikz}
\usepackage{cite}
\usepackage{amsmath,amssymb,amsfonts}
\usepackage{textcomp}
\usepackage{academicons}
\usepackage{hyperref}
\usepackage{xcolor}
\usepackage{acronym}
\usepackage{multirow}
\usepackage{enumitem}
\usepackage{mathabx}
\usepackage{csquotes} 
\usepackage{tabularx,ragged2e,booktabs,caption}
\usepackage{footnote}
\usepackage{threeparttable}
\usetikzlibrary{arrows.meta, positioning}
\usetikzlibrary{shapes.geometric, arrows}
\usepackage{textgreek}
\usepackage{amsmath}
\usepackage{algorithm}
\usepackage{algpseudocode}
\let\oldvec\vec
\renewcommand{\vec}[1]{\oldvec{#1}}

\begin{document}
	\title{Uncertainty Representation in a SOTIF-Related Use Case with Dempster-Shafer Theory for LiDAR Sensor-Based Object Detection}

	\titlerunning{Uncertainty Representation in a SOTIF-Related Use Case with DST}
	
	\author{Milin Patel\inst{1}\orcidID{0000-0002-8357-6018} \and
		Rolf Jung\inst{2}\orcidID{0000-0002-0366-4844}}
	
	\authorrunning{Milin Patel and Rolf Jung}
	
	\institute{Institute for Driver Assistance and Connected Mobility (IFM),
		Benningen, Germany \and
		Kempten University of Applied Sciences, Kempten, Germany \\
		\email{\{milin.patel, rolf.jung\}@hs-kempten.de}}
	
	\maketitle             

\begin{abstract}

Uncertainty in LiDAR sensor-based object detection arises from environmental variability and sensor performance limitations. Representing these uncertainties is essential for ensuring the Safety of the Intended Functionality (SOTIF), which focuses on preventing hazards in automated driving scenarios. This paper presents a systematic approach to identifying, classifying, and representing uncertainties in LiDAR-based object detection within a SOTIF-related scenario. Dempster-Shafer Theory (DST) is employed to construct a Frame of Discernment (FoD) to represent detection outcomes. Conditional Basic Probability Assignments (BPAs) are applied based on dependencies among identified uncertainty sources. Yager's Rule of Combination is used to resolve conflicting evidence from multiple sources, providing a structured framework to evaluate uncertainties' effects on detection accuracy. The study applies variance-based sensitivity analysis (VBSA) to quantify and prioritize uncertainties, detailing their specific impact on detection performance.

	
\end{abstract}
\section{Introduction}
\label{chap:introduction}
The safety of the intended functionality (SOTIF), as defined by ISO 21448 \cite{ISO21448:2022.202206}, focuses on addressing hazards arising from functional insufficiencies in automated driving systems (ADS), particularly those caused by incomplete specifications of the intended functionality. SOTIF emphasizes identifying and mitigating risks associated with these insufficiencies, especially in environments with diverse and unpredictable conditions where operational parameters are not well-defined.

For environmental perception in autonomous driving, ADS rely on a combination of sensors, including cameras, radar, and LiDAR, integrated with machine learning algorithms for object detection and classification. LiDAR sensors are specifically chosen for their capability to provide accurate 3D representations of the environment \cite{Cao.2023}. The process of 3D object detection using LiDAR data includes data acquisition, preprocessing (removing noise and irrelevant points), segmentation (grouping processed points into clusters representing potential objects), feature extraction, and classification. Lastly, tracking monitors detected objects across frames to predict their future positions \cite{Wu.2021}.

However, the performance of LiDAR sensors is significantly affected by environmental conditions, which introduces uncertainty in object detection. These uncertainties arise due to factors like absorption, scattering, and refraction of LiDAR beams, which affect detection range and data quality. Consequently, machine learning models must adapt to handle these uncertainties effectively \cite{TaoPeng.2023}.

Uncertainty can be categorized into two types: aleatoric and epistemic. Aleatoric uncertainty arises from inherent environmental variability, while epistemic uncertainty stems from incomplete knowledge or limitations in the system's model \cite{Gruber.2023, DerKiureghian.2009}. To represent and manage these uncertainties, this paper employs dempster-shafer theory (DST) \cite{Shafer.1976}, a mathematical framework that combines evidence from multiple sources, particularly in scenarios with incomplete or conflicting information \cite{sentz2002combination}.

DST is selected for this research due to its flexibility in representing uncertainty without requiring prior probabilities, which can be difficult to determine or justify \cite{Dezert.2012, Wilson.1993}. Unlike Bayesian methods, which rely on potentially biased prior distributions, DST integrates evidence from multiple sources without enforcing a single probabilistic outcome, making it particularly effective in complex systems where data may be unreliable or contradictory \cite{PeiWang.1994}. Additionally, DST allows for the representation of both uncertainty and ignorance, which is important when evidence is insufficient to fully support any hypothesis \cite{Dezert.2012}. Compared to bayesian probability, fuzzy logic, and possibility theory, DST offers an adaptable approach to uncertainty representation \cite{Zio.2013, Dezert.2012}.

Despite its computational challenges and difficulties in managing high-conflict evidence, DST's method for combining diverse sources of information enables a nuanced representation of uncertainty, justifying its application in this paper for ensuring the reliability of ADS \cite{Wilson.1993, PeiWang.1994, Zio.2013}.

Extended Evidential Networks (EEN), introduced in \cite{Adee.2020}, are used to model and represent aleatoric, epistemic, and ontological uncertainties. The EEN framework integrates plausibility functions from DST into traditional Bayesian networks, capturing these uncertainties. This approach is applied in a SOTIF context through a case study on a perception function in highly automated driving vehicles, highlighting its role in identifying areas for model refinement to improve safety analysis. This work differs by applying DST to LiDAR-based object detection and focusing on developing DST-informed mitigation strategies, particularly through sensitivity analysis.

This paper extends previous research \cite{MilinPatel..2024}, which evaluated the adaptability and performance of deep learning (DL)-based 3D object detection methods using LiDAR data in a SOTIF-related scenario. Building on this foundation, the current work applies DST to specifically address and manage uncertainties in LiDAR-based object detection within the same context. This extension shifts the focus from performance evaluation to a detailed analysis of uncertainty representation, with the objective of developing targeted mitigation strategies informed by DST.

\subsection{Major Contribution}
\label{subchap:contribution}
The main contributions of this paper are summarized as follows: 
\begin{enumerate} 
	
\item [(i)] Presentation of a systematic approach for identifying, classifying, and representing uncertainties in a SOTIF-related scenario involving LiDAR-based object detection. 
	
\item[(ii)] Application of DST to represent uncertainties through a case study on a SOTIF-related scenario involving LiDAR-based object detection.

\end{enumerate}

\subsection{Research Questions} 
This paper aims to address the following research questions:
\begin{enumerate} [label={\upshape\bfseries RQ\arabic*:}, wide = 0pt, leftmargin = 3em]
	
	\item How can Dempster-Shafer Theory (DST) be applied to represent and manage uncertainties in LiDAR-based object detection within a SOTIF-related Use Case? \label{RQ1}
	
	\item How does DST facilitate the quantification and prioritization of identified uncertainties in LiDAR-based object detection, and how do these prioritized uncertainties specifically affect detection accuracy within the defined SOTIF-related scenario? \label{RQ2}
	
\end{enumerate}

\subsection{Structure of the Paper}

Following the introduction, this paper is structured as follows. Chapter \ref{sec:proposed_method} presents the methodology for identifying and representing uncertainty in LiDAR-based object detection within a SOTIF-related context. Chapter \ref{sec:method_application} demonstrates the application of this methodology to a SOTIF-related use case, including sensitivity analysis and mitigation strategy development. Finally, Chapter \ref{chap:conclusion} summarizes the findings and proposes directions for future research.

\section{Proposed Method for Representing Uncertainty in SOTIF-related Use Case}
\label{sec:proposed_method}
\begin{figure}[htbp]
	\centering
	\resizebox{0.8\textwidth}{!}{ 
	\begin{tikzpicture}[node distance=2.5cm, auto]
		\tikzstyle{method} = [rectangle, rounded corners, minimum width=3.5cm, minimum height=1cm, text centered, draw=black, fill=orange!30, text width=5cm, align=center]
		\tikzstyle{arrow} = [thick,->,>=stealth]
		
		\node (usecase) [method] {Define SOTIF-related Use Case for LiDAR-based Object Detection};
		\node (scenario) [method, right of=usecase, node distance=7cm] {Derive Concrete Driving Scenario for Analysis};
		\node (uncertainty) [method, below of=usecase, node distance=3cm] {Determine Sources of Uncertainty};
		\node (states) [method, right of=uncertainty, node distance=7cm] {Define Respective States of Uncertainty Sources};
		\node (dag) [method, below of=uncertainty, node distance=3cm] {Map Dependencies using Directed Acyclic Graph (DAG)};
		\node (dst) [method, right of=dag, node distance=7cm] {Formulate Dempster-Shafer Theory (DST) for Uncertainty Representation};
		
		\draw [arrow] (usecase) -- (scenario);
		\draw [arrow] (scenario) -- (uncertainty);
		\draw [arrow] (uncertainty) -- (states);
		\draw [arrow] (uncertainty) -- (dag);
		\draw [arrow] (states) -- (dst);
		\draw [arrow] (dag) -- (dst);
		
	\end{tikzpicture}
}
	\caption{Workflow for representing uncertainty in a SOTIF-related Use Case}
	\label{fig:workflow_method}
\end{figure}
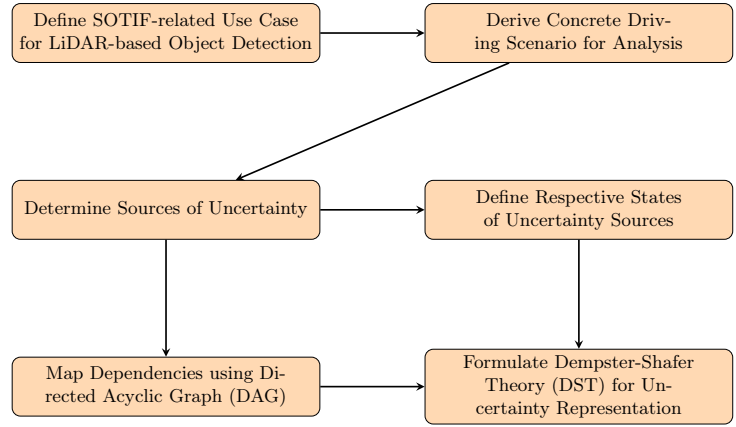

This chapter outlines a methodology for identifying and representing uncertainty in LiDAR-based object detection within a SOTIF-related Use Case. The workflow, illustrated in Figure \ref{fig:workflow_method}, includes defining the Use Case, categorizing sources of uncertainty, and applying DST to represent these uncertainties. The workflow follows a parallel approach, where defining uncertainty states and mapping dependencies occur simultaneously, ensuring all influencing factors are considered together.

\subsection{Description of SOTIF-Use Case and Scenario}
\label{subchap:sotif_description}
\begin{figure}[!h]
	\centering 
	\includegraphics[width=\linewidth]{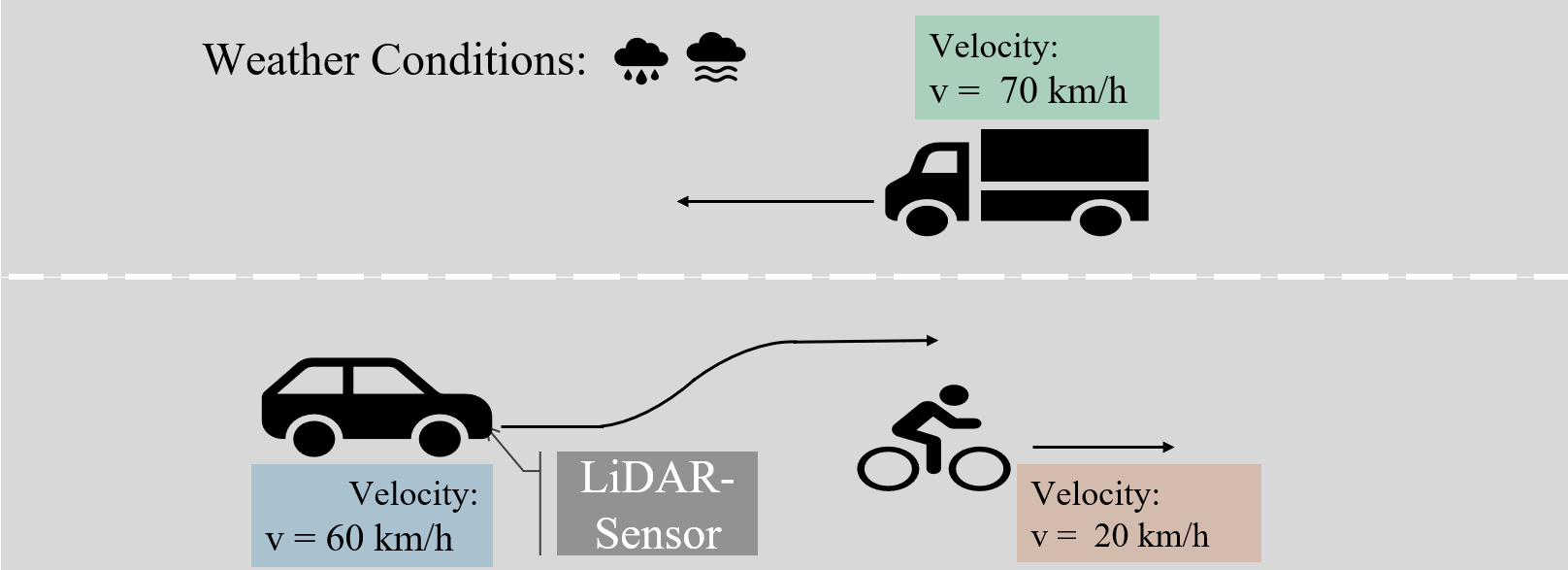}
	\caption{SOTIF-related Use Case: Ego-Vehicle interacting with a cyclist and oncoming truck in adverse weather.}
	\label{fig:usecase}
\end{figure}

This subchapter describes the SOTIF-related Use Case and the derived scenario, which involves a LiDAR-equipped vehicle navigating a two-lane country road under adverse weather conditions. Figure \ref{fig:usecase} depicts the scenario, where the Ego-Vehicle must avoid a cyclist while managing oncoming traffic and challenging weather conditions.

The key variables defining the Use Case, including road type, weather conditions, and road users, are detailed in Table \ref{table:usecase_description}.
\begin{table}[h!]
	\centering
	\caption{Detailed Description of SOTIF-Related Use Case Variables}
	\begin{adjustbox}{max width=\textwidth}
		\begin{threeparttable}
			\begin{tabular}{| m{5cm} | m{10cm} |}
				\hline
				\multicolumn{2}{|c|}{\textbf{Operational Design Domain (ODD) Taxonomy}} \\
				\hline
				Permanent-Regional Variable\tnote{*} & 
				\begin{itemize}
					\item Roadway Type $\rightarrow$ Traffic way - \textit{Two-way, Divided}
					\item Roadway Surface and Features Type $\rightarrow$ Lane Type – \textit{Single Lane, Asphalt}
				\end{itemize} \\
				\hline
				Permanent-Local Variable\tnote{*} & 
				\begin{itemize}
					\item Road Geometry $\rightarrow$ Alignment – \textit{Straight}
					\item Lane Type - \textit{Narrow Lane}
				\end{itemize} \\
				\hline
				Compounding Event or Condition Scenario Variable\tnote{*} & 
				\begin{itemize}
					\item Weather $\rightarrow$ Particulate Matter $\rightarrow$ \textit{Fog}
					\item Weather $\rightarrow$ Precipitation – \textit{Rain}
					\item Light Conditions $\rightarrow$ Ambient Light – \textit{Daylight}
				\end{itemize} \\
				\hline
				Non-typical Event and Condition Scenario Variable\tnote{*} & 
				\begin{itemize}
					\item Roadway Users $\rightarrow$ Non-vehicle Permitted on Roadway - \textit{Bicyclist}
				\end{itemize} \\
				\hline
			\end{tabular}
			\begin{tablenotes}
				\item The variables are categorized based on the study in \cite{Becker.2020}.
			\end{tablenotes}
		\end{threeparttable}
	\end{adjustbox}
	\label{table:usecase_description}
\end{table}

To systematically represent SOTIF-related aspects, Table \ref{table:scenario_description} outlines how the scenario is derived from the defined Use Case, focusing on triggering conditions, performance insufficiencies, and potential hazardous behaviors.

\begin{table}[h!]
	\centering
	\caption{Systematic Description of SOTIF-Related Use Case and Derived Scenario}
	\begin{adjustbox}{max width=\textwidth}
		\begin{threeparttable}
			\begin{tabular}{| m{5cm} | m{10cm} |}
				\hline
				\textbf{Use Case} & Operating on a two-way country road \\
				\hline
				\multicolumn{2}{|c|}{\textbf{Scenario Description}} \\
				\hline
				\multicolumn{2}{|m{15cm}|}{
					The scenario involves a two-lane country road with one lane for each direction and no dedicated cyclist lane. The Ego-Vehicle, equipped with a LiDAR sensor, travels at 60 km/h in the right lane. Ahead, a cyclist moves at 20 km/h on the right side of the road, while a truck approaches in the opposite lane at 70 km/h. Adverse weather conditions, including rain and fog patches, challenge the LiDAR sensor's performance. The Ego-Vehicle detects the cyclist and the oncoming truck, then plans to overtake the cyclist by temporarily moving into the adjacent lane.
				} \\ \hline
				\textbf{Triggering Condition (TC)} & 
				\begin{itemize}
					\item Fog causing noise in LiDAR point cloud data.
					\item Wet road surface increasing braking distance.
				\end{itemize} \\
				\hline
				\textbf{Performance Insufficiency} & 
				\begin{itemize}
					\item LiDAR sensor performance degrades in adverse weather conditions (rain and fog).
					\item Insufficient training data for the deep learning model to handle foggy or rainy conditions.
				\end{itemize} \\
				\hline
				\textbf{Potential Hazardous Behavior} & 
				\begin{itemize}
					\item Incorrect estimation of the cyclist's position, leading to a potential collision.
					\item Inaccurate perception of the oncoming truck, leading to a potential head-on collision.
				\end{itemize} \\
				\hline
			\end{tabular}
		
		\end{threeparttable}
	\end{adjustbox}
	\label{table:scenario_description}
\end{table}

\subsection{Sources of Uncertainty in SOTIF-Related Use Case for LiDAR Sensor-Based Object Detection}
\label{subsec:sources_uncertainty}
This subchapter identifies and categorizes the sources of uncertainty in LiDAR-based object detection, distinguishing between aleatoric and epistemic uncertainties, as shown in Table \ref{table:categorization_uncertainty_sources}. Aleatoric uncertainties arise from environmental factors, including rain intensity, fog density, and road surface conditions, which directly impact the performance of the LiDAR sensor. Epistemic uncertainties result from system limitations, encompassing the sensor's performance under specific conditions, the presence of noise in the data, and the adequacy of the DL model’s training, which influence the accuracy of object detection and classification.
\begin{table}[h!]
	\centering
	\caption{Categorization of Sources of Uncertainty in LiDAR-Based Object Detection}
	\begin{adjustbox}{max width=\textwidth}
			\begin{threeparttable}
		\begin{tabular}{|m{5cm}|m{3cm}|m{8cm}|}
			\hline
			\textbf{Uncertainty Source} & \textbf{Category} & \textbf{Interdependencies} \\ \hline
			Rain Intensity & Aleatoric & Variability in rain intensity affects road wetness, surface reflectivity, and LiDAR signal scattering. \\ \hline
			Fog Density & Aleatoric & Changes in fog density cause LiDAR signal scattering and noise in data. \\ \hline
			Wet Road Conditions & Aleatoric & Moisture levels on the road, influenced by rain and fog, affect reflection variability and surface characteristics. \\ \hline
			LiDAR Sensor Performance & Aleatoric \& Epistemic & Performance is influenced by environmental factors (aleatoric) and sensor limitations (epistemic). \\ \hline
			Noise in LiDAR Data & Aleatoric & Environmental factors like rain and fog introduce noise into LiDAR data, reducing accuracy. \\ \hline
			Scattering of LiDAR Signals & Aleatoric & Surface type, fog, and rain cause scattering, reducing signal clarity. \\ \hline
			Reflection Variability & Aleatoric & Reflection variability arises from wet road conditions and surface type characteristics. \\ \hline
			Object Proximity & Aleatoric & Variability in detecting objects based on distance and speed relative to the sensor. \\ \hline
			Surface Type & Aleatoric & Different surface characteristics (absorption, transmission, reflection) impact how LiDAR signals behave. \\ \hline
			Deep Learning Model Training Quality & Epistemic & The diversity and quality of training data impact model generalization and accuracy. \\ \hline
			3D Object Detection Accuracy & Epistemic & Uncertainty arises from sensor data limitations and model inaccuracies in detecting objects. \\ \hline
		\end{tabular}
	 \begin{tablenotes}
		\item The sources of uncertainty listed in this table are adapted from SOTIF scenario variables \cite{Becker.2020}, and LiDAR sensor model and 3D object detection deep learning model parameters as described in \cite{Dosovitskiy.2017} and \cite{Prince.2023}.
	\end{tablenotes}
			\end{threeparttable}
	\end{adjustbox}
	\label{table:categorization_uncertainty_sources}
\end{table}
\subsection{Representing Dependencies Between Uncertainty Sources}
\label{subsec:dependencies_uncertainty}
\begin{figure}[h!]
	\centering 
	\includegraphics[width=0.8\textwidth]{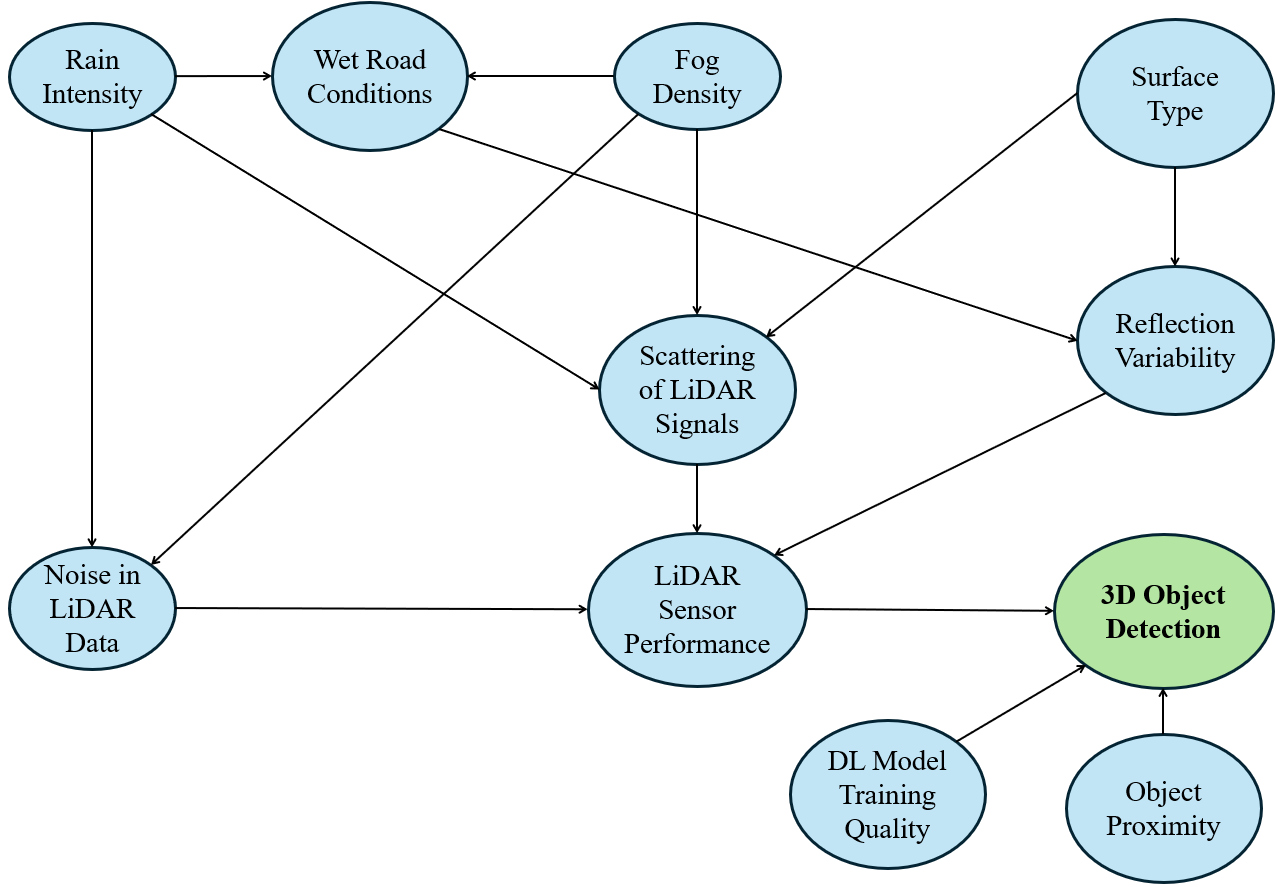}
	\caption{DAG representing dependencies among uncertainty sources in a SOTIF-related Use Case for LiDAR-based object detection}
	\label{fig:dag}
\end{figure}

A directed acyclic graph (DAG), shown in Figure \ref{fig:dag}, is used to map the dependencies between various sources of uncertainty in LiDAR-based object detection. This representation demonstrates how environmental and system factors interact to influence detection accuracy. Each node in the DAG corresponds to a specific source of uncertainty, while the edges indicate the conditional relationships between these sources, showing how they collectively impact the system's overall performance and decision-making processes \cite{Hewawasam.2007}.

As depicted in Figure \ref{fig:dag}, the dependencies show that rain intensity and fog density directly affect wet road conditions, which subsequently influence both reflection variability and LiDAR sensor performance. Surface type also affects the scattering of LiDAR signals and reflection variability. These factors collectively influence LiDAR sensor performance, directly affecting the accuracy of 3D object detection. Additionally, DL model training quality and object proximity impact object detection performance, determining the system's ability to accurately identify and classify objects within the environment.

The identified uncertainties are classified into distinct states, as summarized in Table \ref{table:uncertainty_states}. These states represent the specific levels or conditions under which each source of uncertainty may occur. In the following chapter, these uncertainties will be analyzed using DST by assigning conditional Basic Probability Assignments (BPAs) to each source, according to its respective state.

\begin{table}[h!]
	\centering
	\caption{State Categories of Uncertainty Sources}
	\begin{adjustbox}{max width=\textwidth}
		\begin{tabular}{| m{5cm} | m{10cm} |}
			\hline
			\textbf{State Category} & \textbf{Uncertainty Sources} \\ \hline
			Low, Medium, High & Rain Intensity, Fog Density, Noise in LiDAR Data, Scattering of LiDAR Signals, Reflection Variability \\ \hline
			Dry, Moist, Saturated & Wet Road Conditions \\ \hline
			Good, Moderate, Poor & LiDAR Sensor Performance, Deep Learning Model Training Quality \\ \hline
			Close, Medium, Far & Object Proximity \\ \hline
			Absorption, Transmission, Reflection & Surface Type \\ \hline
		\end{tabular}
	\end{adjustbox}
	\label{table:uncertainty_states}
\end{table}

\subsection{Formulating Dempster-Shafer Theory}
\label{subsec:dst_formulation}
In this study, DST is applied to represent and manage uncertainty in LiDAR-based object detection. The main components of DST include the Frame of Discernment (FoD), BPA, and the belief ($Bel$) and plausibility ($Pl$) functions. These components work together to provide a structured framework for uncertainty representation \cite{Rakowsky.2007, Campos.2003}.

\textbf{Frame of Discernment (FoD)}: The FoD, denoted as $\Theta$, represents the set of all possible outcomes or hypotheses within the system. Each element $\theta_i$ corresponds to a potential outcome of the LiDAR-based object detection process, and the FoD is defined as:
\begin{equation}
	\Theta = \{\theta_1, \theta_2, \theta_3, \ldots, \theta_n\}
\end{equation}

\textbf{Basic Probability Assignment (BPA)}: The BPA, $m(A)$, assigns a measure of belief to each subset $A \subseteq \Theta$, indicating the degree of evidence that supports $A$. The BPA must satisfy:
\begin{equation}
	m(\emptyset) = 0 \quad \text{and} \quad \sum_{A \subseteq \Theta} m(A) = 1
\end{equation}
In this study, BPAs are derived from simulation data, capturing both aleatory and epistemic uncertainties.

\textbf{Belief and Plausibility Functions}: The belief function, $Bel(A)$, and the plausibility function, $Pl(A)$, provide lower and upper bounds, respectively, for the probability of $A$. The belief function is defined as:
\begin{equation}
	Bel(A) = \sum_{B \subseteq A} m(B)
\end{equation}
This represents the minimum belief committed to $A$ based on the available evidence. The plausibility function is defined as:
\begin{equation}
	Pl(A) = \sum_{B \cap A \neq \emptyset} m(B)
\end{equation}
which accounts for all evidence that does not contradict $A$. The interval $[Bel(A), Pl(A)]$ represents the range within which the true probability of $A$ lies, accommodating both certainty and uncertainty \cite{Klir.1995}.

\textbf{Dempster's Rule of Combination}: Dempster's rule is used to combine evidence from multiple sources. The combined BPA for any subset $C \subseteq \Theta$, denoted as $m_{12}(C)$, is calculated as:
\begin{equation}
	m_{12}(C) = \frac{1}{1 - K} \sum_{A \cap B = C} m_1(A) \times m_2(B)
\end{equation}
where $K$ is the conflict coefficient, quantifying the degree of conflict between evidence sources:
\begin{equation}
	K = \sum_{A \cap B = \emptyset} m_1(A) \times m_2(B)
\end{equation}
Dempster's rule ensures that conflicting evidence is appropriately weighted, with $m_{12}(C)$ representing the combined belief in $C$ based on both sources \cite{A.P.Dempster.1967}.

\textbf{Yager's Modified Rule of Combination}: In cases of significant conflict between evidence sources, Yager's modification of Dempster's rule is applied. This approach redistributes conflicting belief to the universal set $\Theta$, with the modified belief assigned to $\Theta$ as:
\begin{equation}
	q(\Theta) = m_1(\Theta) \times m_2(\Theta) + \sum_{A \cap B = \emptyset} m_1(A) \times m_2(B)
\end{equation}
This method retains and addresses conflicting information without normalization, making it particularly suitable for complex systems like LiDAR-based object detection, where multiple forms of evidence must be reconciled \cite{sentz2002combination}. In this study, Yager's rule is employed to manage the high levels of conflict present in the evidence sources.

\section{Method Application on SOTIF-related Use Case}
\label{sec:method_application}
\begin{figure}[htbp]
	\centering
	\resizebox{0.8\textwidth}{!}{
		\begin{tikzpicture}[node distance=2.5cm, auto]
			
			\tikzstyle{implement} = [rectangle, rounded corners, minimum width=3.5cm, minimum height=1cm, text centered, draw=black, fill=green!30, text width=5cm, align=center]
			\tikzstyle{arrow} = [thick,->,>=stealth]
			
			\node (fod) [implement] {Define Frame of Discernment (FoD)};
			\node (bpa) [implement, right of=fod, node distance=7cm] {Assign Conditional Basic Probability Assignments (BPAs)};
			\node (combine) [implement, below of=fod, node distance=3cm] {Combine Conflicting Evidence Using Yager's Rule};
			\node (belief) [implement, right of=combine, node distance=7cm] {Calculate Belief and Plausibility};
			\node (categorization) [implement, below of=combine, node distance=3cm] {Categorize Uncertainty Sources by Impact Level};
			\node (sensitivity) [implement, below of=belief, node distance=3cm] {Conduct Variance-Based Sensitivity Analysis};
			\node (mitigation) [implement, below of=sensitivity, node distance=3cm] {Define Mitigation Measures};
			
			\draw [arrow] (fod) -- (bpa);
			\draw [arrow] (fod) -- (combine);
			\draw [arrow] (bpa) -- (combine);
			\draw [arrow] (combine) -- (belief);
			\draw [arrow] (belief) -- (categorization);
			\draw [arrow] (categorization) -- (sensitivity);
			\draw [arrow] (sensitivity) -- (mitigation);
		\end{tikzpicture}
	}
	\caption{Workflow for applying DST in a SOTIF-related Use Case}
	\label{fig:workflow}
\end{figure}
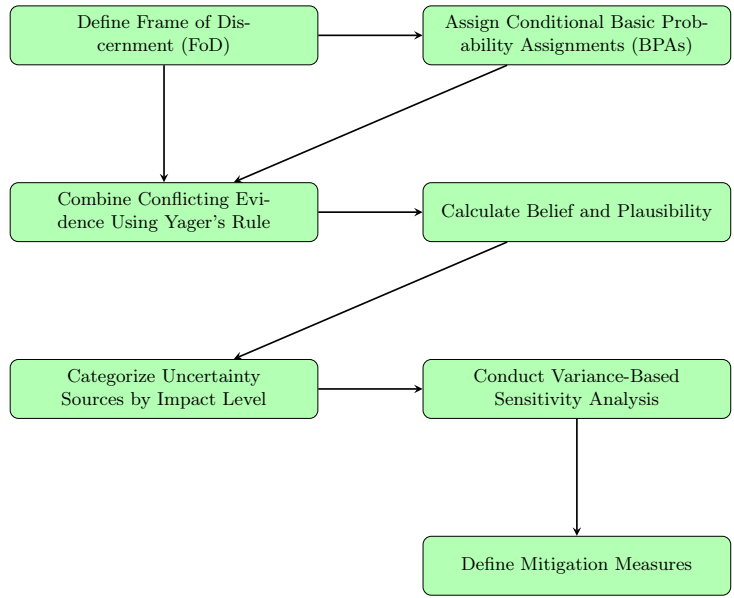

This chapter applies DST to a SOTIF-related Use Case in LiDAR-based object detection. The workflow in Figure \ref{fig:workflow} outlines the process of defining the FoD, categorizing uncertainty sources based on their impact on detection accuracy, performing a variance-based sensitivity analysis, and suggesting mitigation measures for the most significant uncertainties. The FoD defines possible detection outcomes, and BPAs assign probabilities based on uncertainty sources. Both inputs are necessary for the combination process, enabling conflict resolution through Yager's Rule.

This parallel structure allows the combination of evidence to consider both the hypotheses and their assigned probabilities simultaneously, enabling conflict resolution through Yager's Rule. 
\subsection{Defining the Frame of Discernment (FoD)}
\label{subsec:defining_fod}
The Frame of Discernment (FoD) for this Use Case includes four possible outcomes relevant to LiDAR-based object detection:
\begin{itemize}
	\item $\theta_1$: Detection of a Cyclist
	\item $\theta_2$: Detection of a Truck
	\item $\theta_3$: No Detection, but object present (False Negative)
	\item $\theta_4$: Incorrect Detection, but no object present (False Positive)
\end{itemize}
These outcomes, represented as $\Theta = \{\theta_1, \theta_2, \theta_3, \theta_4\}$, form the basis for analyzing how uncertainties impact the performance of the LiDAR-based object detection system. A false negative refers to a missed detection when an object is actually present, while a false positive refers to an incorrect detection when no object is present.

\subsection{Assigning Conditional Basic Probability Assignments (BPAs)}
\label{subsec:assigning_bpas}
\begin{figure}[h!]
	\centering 
	\includegraphics[width=\linewidth]{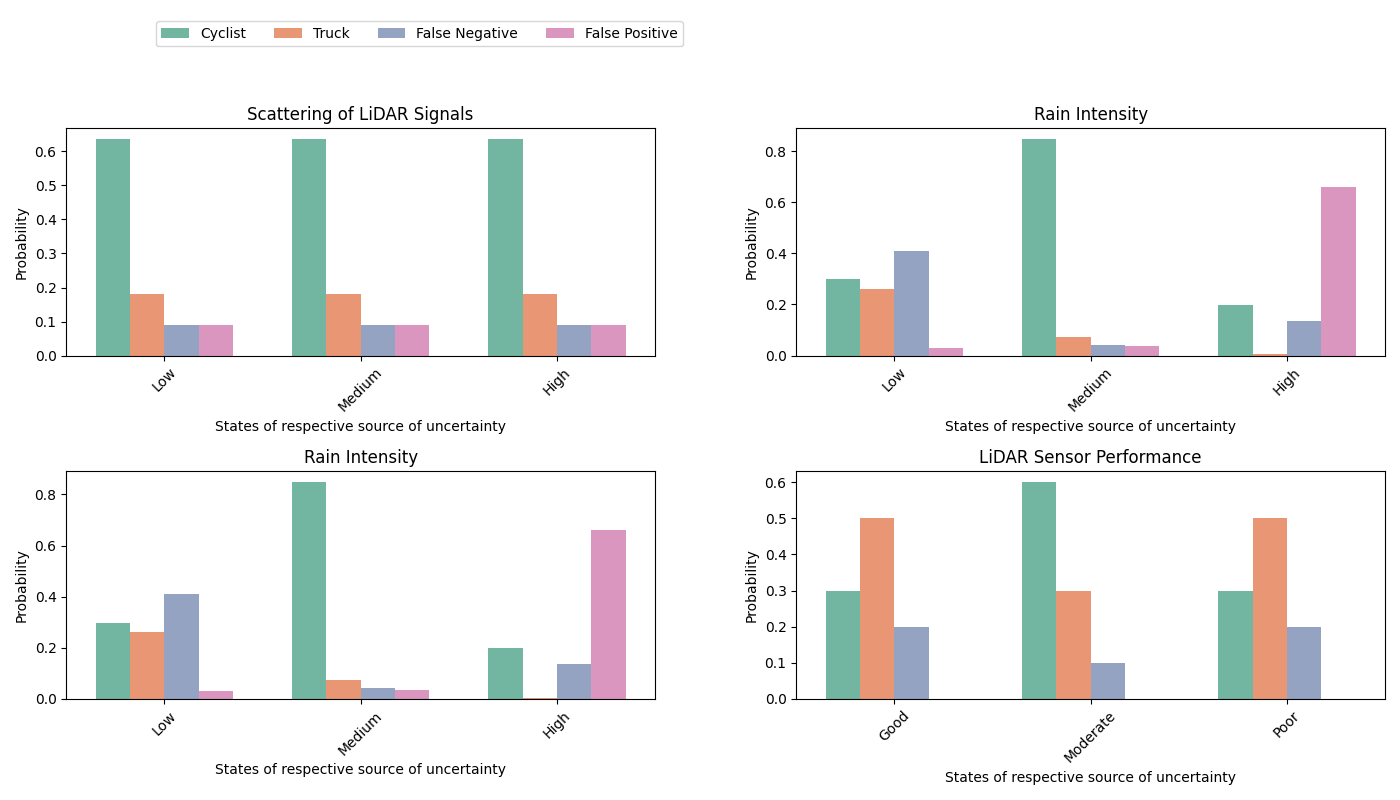}
	\caption{Conditional Probability distributions for detection outcomes across various sources of uncertainty}
	\label{fig:bpa_results}
\end{figure}

Conditional BPAs are assigned based on the dependencies among various sources of uncertainty, as illustrated in Figure \ref{fig:bpa_results}. The uncertainty sources, denoted as $S = \{S_1, S_2, \ldots, S_n\}$ (Table \ref{table:categorization_uncertainty_sources}), have corresponding states $X_s$ as shown in Table \ref{table:uncertainty_states}. For each detection outcome $\Theta = \{\theta_1, \theta_2, \theta_3, \theta_4\}$, conditional BPAs $m(x)$ are generated based on the dependencies of each source state $x \in X_s$ and are normalized to ensure that $\sum_{A \subseteq \Theta} m(A) = 1$.

Figure \ref{fig:bpa_results} presents selected uncertainty sources to provide a focused evaluation, as including all uncertainties would lead to repetitive information without adding new insights. The results highlight the conditional dependencies between uncertainties, revealing how environmental conditions influence object detection outcomes.

Rain intensity leads to a significant increase in False Positives as intensity rises, reflecting reduced sensor performance in adverse weather such as heavy rain. In contrast, fog density shows stable false positive probabilities across states but greater variability in detecting cyclists and trucks, indicating reduced sensor reliability under foggy conditions. Similarly, scattering of LiDAR signals correlates with increased False Negatives at higher scattering levels, suggesting compromised detection accuracy for objects like cyclists and trucks when environmental factors distort the signal.

\subsection{Combining BPAs Using Yager's Rule of Combination and Calculating Belief and Plausibility}
\label{subsec:combining_bpas}
Yager's Rule of Combination was applied to integrate BPAs from different sources of uncertainty, particularly when evidence is conflicting. Conflicting evidence occurs when different sources provide varying levels of belief in certain outcomes, making it challenging to combine them directly without bias. Yager's rule redistributes the conflicting mass to the universal set $\Theta$, ensuring that no single outcome is disproportionately influenced by conflicting evidence.
\begin{figure}[h!]
	\centering 
	\includegraphics[width=\linewidth]{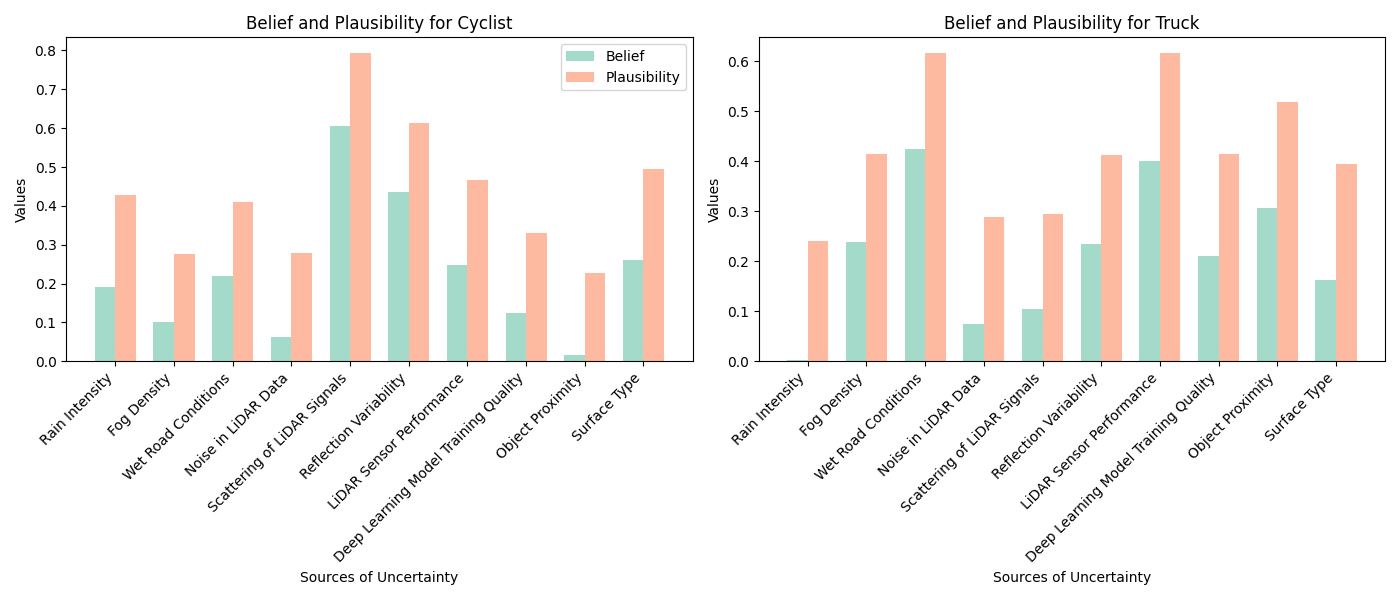}
	\caption{$Bel$ and $Pl$ for detection outcomes across various sources of uncertainty}
	\label{fig:belief_plausibility}
\end{figure}
Figure \ref{fig:belief_plausibility} demonstrates how different uncertainty sources affect the belief and plausibility values for each detection outcome. A higher plausibility value compared to belief indicates that the outcome is possible across a broader range of scenarios, although with less certainty. 

For cyclist detection, reflection variability and rain intensity show higher plausibility values compared to belief, suggesting the system detects cyclists under specific conditions but with reduced confidence in accuracy. For truck detection, reflection variability and LiDAR sensor performance play a significant role in reducing detection confidence under adverse conditions. While similar patterns are observed in other detection outcomes, only cyclist and truck detections are shown here to avoid redundant information.

\subsection{Impact Levels of Uncertainty Sources}
\label{subsec:categorization_uncertainty}
Uncertainty sources were categorized by calculating the difference between their plausibility and belief values, represented as $U(A) = Pl(A) - Bel(A)$ for each subset $A \subseteq \Theta$. This difference quantifies the uncertainty associated with each source. The uncertainty levels were classified into three categories: Low, Moderate, and High, based on thresholds derived from the distribution of uncertainty values.

The classification thresholds were set using the 25\textsuperscript{th} and 75\textsuperscript{th} percentiles of the maximum uncertainty values across all sources, corresponding to $\tau_1 = 0.2$ and $\tau_2 = 0.5$. Sources with $U(A) < \tau_1$ were categorized as Low Impact, those with $\tau_1 \leq U(A) < \tau_2$ as Moderate Impact, and those with $U(A) \geq \tau_2$ as High Impact. These thresholds effectively segment the uncertainty distribution, as shown in Table \ref{tab:summary_sources_of_uncertainty}.
\begin{table}[h!]
	\centering
	\caption{Impact Levels of Uncertainty Sources on Detection Variability}
	\label{tab:summary_sources_of_uncertainty}
	\begin{adjustbox}{max width=\textwidth}
		\begin{tabular}{| m{3cm} | m{12cm} |} \hline
			\textbf{Impact Level} & \textbf{Sources of Uncertainty} \\ \hline
			High Impact & Rain Intensity, LiDAR Sensor Performance, Surface Type \\ \hline
			Moderate Impact & Wet Road Conditions, Noise in LiDAR Data, Deep Learning Model Training Quality, Object Proximity \\ \hline
			Low Impact & Fog Density, Scattering of LiDAR Signals, Reflection Variability \\ \hline
		\end{tabular}
	\end{adjustbox}
\end{table}

The impact levels in Table \ref{tab:summary_sources_of_uncertainty} reflect the overall effect of each uncertainty source on detection performance, while Table \ref{table:uncertainty_states} lists the specific states (e.g., Low, Medium, High) under which these uncertainties occur. The states serve as inputs to determine their contribution to the overall impact levels.

\subsection{Variance-Based Sensitivity Analysis (VBSA)}
\label{subsec:sensitivity_analysis}
After categorizing uncertainties by their impact levels, it was essential to quantify their contributions to detection variability. VBSA was applied to evaluate how variations in input uncertainties influence overall detection performance, particularly in systems with interdependent factors \cite{Liu.2024}.

The analysis focuses on uncertainty sources to assess their impact on detection accuracy under different environmental conditions. VBSA systematically allocates the total variance in performance across these uncertainty sources, providing insights into how each source interacts with others \cite{D.Shahsavani.2011}. This method identifies the sources with the most significant impact on performance variability \cite{JimW.Hall.2006}, which is essential for managing uncertainty and improving system reliability.

\begin{equation}
	\text{Var}_{Bel}(S_i) = \text{Var}(Bel(S_i)), \quad \text{Var}_{Pl}(S_i) = \text{Var}(Pl(S_i))
\end{equation}

The variance for each uncertainty source $S_i$ is calculated to determine how belief and plausibility values fluctuate under different conditions. Comparing these variances helps identify which sources contribute most to system variability.

\begin{figure}[h!]
	\centering 
	\includegraphics[width=\linewidth]{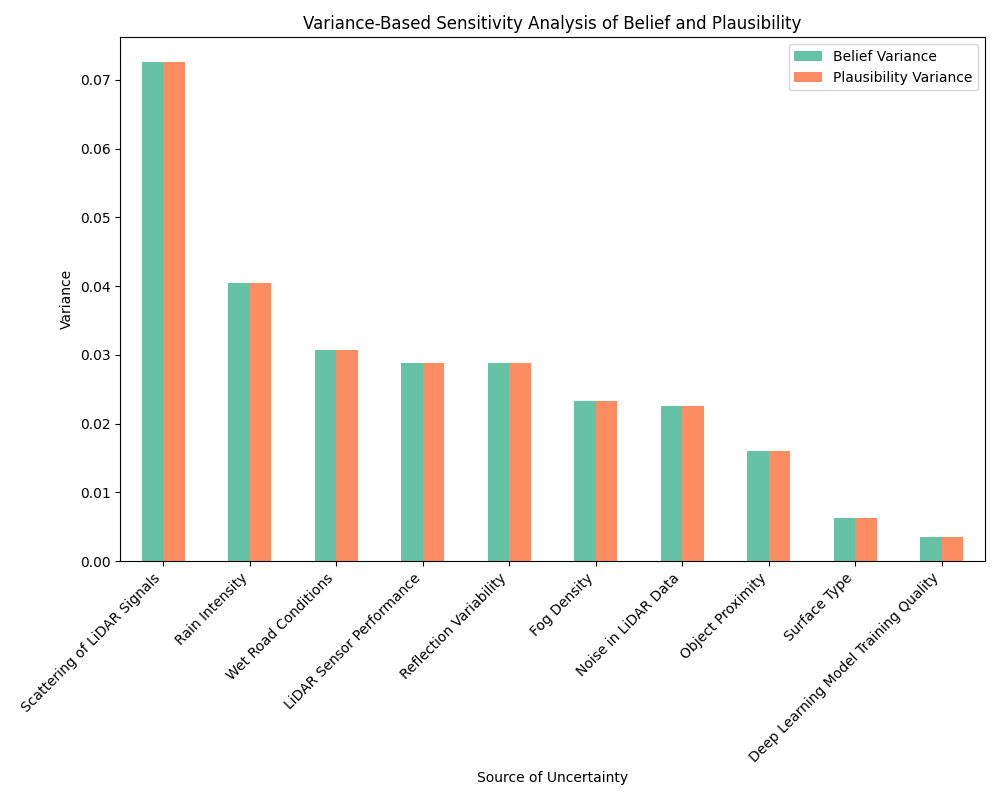}
	\caption{Variance-Based sensitivity analysis of $Bel$ and $Pl$}
	\label{fig:sensitivity_analysis}
\end{figure}

As shown in Figure \ref{fig:sensitivity_analysis}, the VBSA results shows that scattering of LiDAR signals and rain intensity generate the highest variance in detection performance, particularly under challenging environmental conditions. Wet road conditions and LiDAR sensor performance also contribute significantly to performance variability.

\subsection{Mitigation Measures to Address Key Uncertainties}
\label{subsec:mitigation_measures}
Based on the findings from VBSA, mitigation measures are suggested to address uncertainties that significantly impact the performance of the LiDAR-based object detection system. The recommendations focus on reducing detection variability and enhancing system robustness across different operational environments:
\begin{enumerate}[label=(\alph*)]
	\item Develop adaptive algorithms for LiDAR-based object detection to adjust to varying weather conditions, including rain intensity and fog density, improving detection accuracy in challenging environments.
	\item Apply noise reduction techniques to minimize the impact of environmental interference on LiDAR data, enhancing detection performance in adverse conditions.
	\item Expand training datasets with simulated data from extreme weather conditions to improve the model's generalization and maintain accuracy across different real-world scenarios.
\end{enumerate}
These measures directly address the most significant uncertainties identified through the DST framework and VBSA.
\section{Conclusion and Future Research Directions}
\label{chap:conclusion}
This paper investigates the application of Dempster-Shafer Theory (DST) to address uncertainties in LiDAR-based object detection within a SOTIF-related Use Case. To answer the first research question (\ref{RQ1}), the study defines a SOTIF-related Use Case (Chapter \ref{subchap:sotif_description}), identifies relevant sources of uncertainty (Table \ref{table:categorization_uncertainty_sources}), and applies DST to model these uncertainties. The Frame of Discernment (FoD) models detection outcomes, and BPAs are calculated based on the available evidence (Chapter \ref{subsec:dst_formulation}). Yager's Rule of Combination is used to resolve conflicting information from different sources (Chapter \ref{subsec:combining_bpas}), providing a structured and objective representation of uncertainty.

To address the second research question (\ref{RQ2}), the study quantifies and prioritizes uncertainties based on their impact on detection performance. Using belief and plausibility functions, the uncertainties are categorized by their influence on detection accuracy (Chapter \ref{subsec:dependencies_uncertainty}). The VBSA further quantifies the contribution of each uncertainty source (Chapter \ref{subsec:sensitivity_analysis}), with environmental factors such as rain intensity, surface type, and LiDAR sensor performance having the most significant effect on detection variability (Figure \ref{fig:sensitivity_analysis}).

Mitigation measures are proposed to address the identified uncertainties, focusing on improving LiDAR sensor performance under adverse weather conditions and refining training datasets to include extreme weather scenarios. These recommendations, discussed in Chapter \ref{subsec:mitigation_measures}, aim to improve system reliability by addressing critical environmental and sensor limitations.

Several limitations are acknowledged in this study. First, the subjectivity in assigning BPAs may introduce bias. Second, the static analysis does not capture temporal dynamics, which are crucial for real-time decision-making in dynamic environments. Additionally, the conclusions are sensitive to the initial assumptions and the choice of combination rules in DST.

Future research could extend this static analysis into a dynamic framework, tracking uncertainty propagation over time as the vehicle interacts with its environment. Modeling the evolution of uncertainties in sensor performance, environmental conditions, and object proximity would improve real-time decision-making and risk assessment in Automated Driving Systems (ADS). Further exploration of uncertainty propagation in different driving scenarios and integrating DST with temporal models, such as Hidden Markov Models (HMMs) or Bayesian Networks, could reveal additional insights into system behavior. Simulation-based studies could also validate these uncertainty models by comparing their predictions with real-world data, leading to more effective uncertainty management techniques.

\begin{credits}
\subsubsection{\ackname} This research was funded by the Institute for Driver Assistance and Connected Mobility (IFM) at Kempten University of Applied Sciences. The IFM specializes in the development and validation of driver assistance systems and connected mobility, focusing on functional safety, cybersecurity, and testing methodologies. I extend my gratitude to my team at IFM for their valuable support and contributions in the areas of functional safety and cybersecurity. 
\end{credits}
  
\bibliographystyle{IEEEtran}
\bibliography{main_submit}

\end{document}